\documentclass[letterpaper, 10 pt, conference]{ieeeconf}
\IEEEoverridecommandlockouts 
\usepackage{graphics} 
\usepackage{epsfig} 
\usepackage{amsmath} 
\usepackage{amssymb, bm} 
\usepackage[ruled,vlined,linesnumbered]{algorithm2e}
\usepackage[dvipsnames]{xcolor} %

\DeclareMathOperator*{\argmin}{argmin} 

\title{\LARGE \bf
Planning for Safe Abortable Overtaking Maneuvers in \linebreak Autonomous Driving  
}

\author{Jiyo Palatti$^{1}$, Andrei Aksjonov$^{1}$, Gokhan Alcan$^{1}$, and Ville Kyrki$^{1}$, \IEEEmembership{Senior Member, IEEE}%

\thanks{This work was supported by Academy of Finland grant 328399.}
\thanks{$^{1}$The authors are with the Intelligent Robotics group, Department of Electrical Engineering and Automation, Aalto University, 02150 Espoo, Finland \{ {\tt\small jiyo.palatti, andrei.aksjonov, gokhan.alcan, ville.kyrki} \} {\tt\small @aalto.fi}
}}

\begin{document}

\maketitle
\thispagestyle{empty}
\pagestyle{empty}

\begin{abstract}
Overtaking is one of the most challenging tasks in driving, and the current solutions to autonomous overtaking are limited to simple and static scenarios. 
In this paper, we present a method for behaviour and trajectory planning for safe autonomous overtaking. 
The proposed method optimizes the trajectory by simultaneously enforcing safety and minimizing intrusion onto the adjacent lane. Furthermore, the method allows the overtaking to be aborted, enabling the autonomous vehicle to merge back in the lane, if safety is compromised, because of e.g. traffic in opposing direction appearing during the maneuver execution. A finite state machine is used to select an appropriate maneuver at each time, and a combination of safe and reachable sets is used to iteratively generate intermediate reference targets based on the current maneuver. A nonlinear model predictive controller then plans dynamically feasible and collision-free trajectories to these intermediate reference targets.
Simulation experiments demonstrate that the combination of intermediate reference generation and model predictive control is able to handle multiple behaviors, including following a lead vehicle, overtaking and aborting the overtake, within a single framework. 
\end{abstract}

\section{Introduction}
An Autonomous Vehicle (AV) must be able to execute several complex driving maneuvers, such as lane keeping, lane changing, and overtaking, that are involved in typical driving situations. 
The ability to overtake is essential to increase road capacity and level of service, especially on single and two lane roads \cite{farahWhenDriversAbort2016}.
However, overtaking on such roads is one of the most challenging maneuvers, because it requires the vehicle to drive on the path of potential oncoming traffic for significant periods of time, often at high speeds.  

Existing methods consider usually simple scenarios, where a Leading Vehicle (LV) to be overtaken is static or moving at a constant velocity without oncoming vehicles \cite{dixitTrajectoryPlanningTracking2018}. In these settings, overtaking can be understood as two successive lane change maneuvers. However, when oncoming traffic exists, this is not sufficient.   

\begin{figure}[!t]
    \centering
    \includegraphics[width=\linewidth]{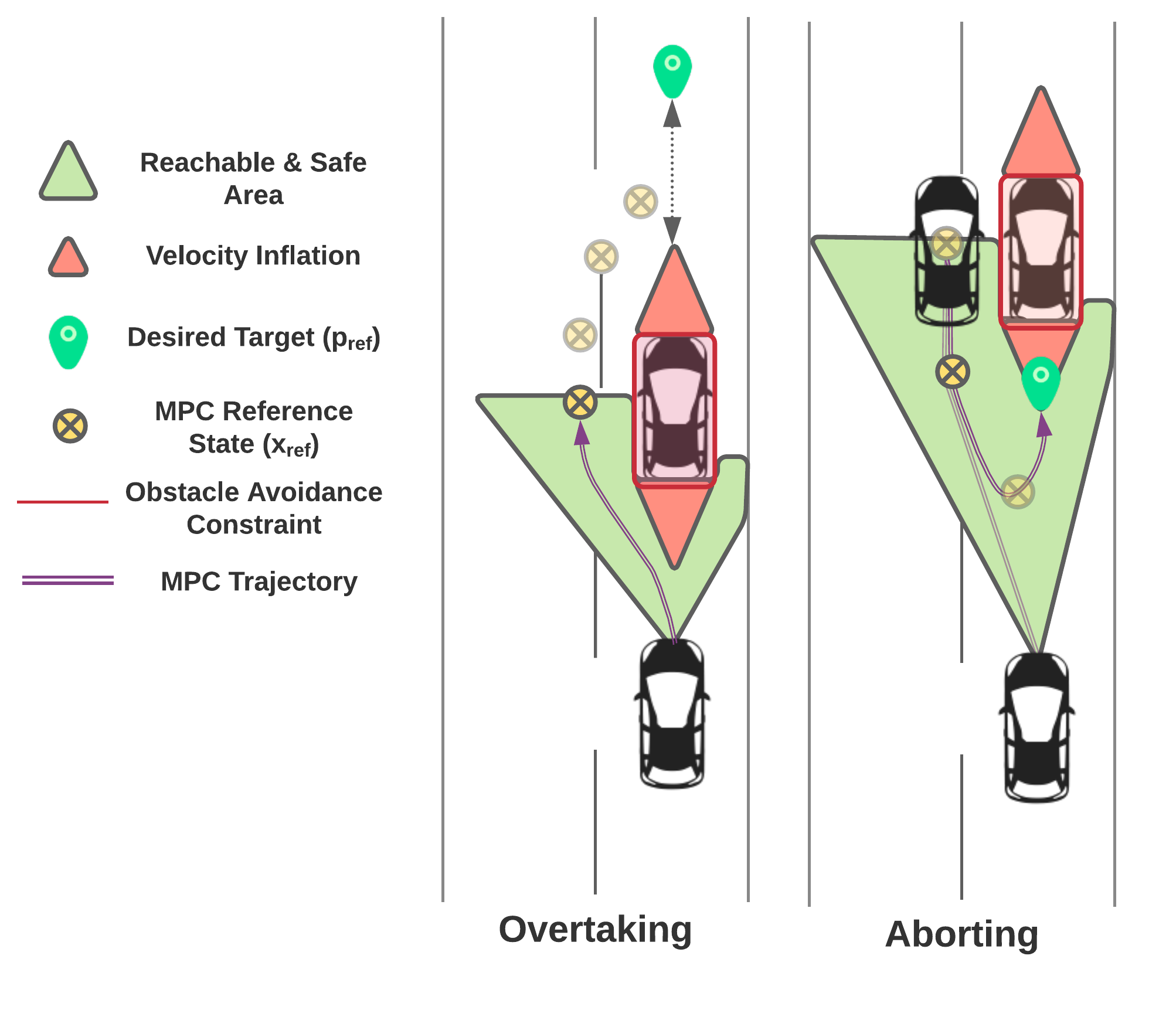}
    \caption{Proposed autonomous overtaking method minimizes intrusion on adjacent lane while ensuring safe clearance during overtake and allows aborting the overtake if needed. 
    \label{intermediaterefselection}}
\end{figure}

We address the issue of increasing the safety of overtaking under potential oncoming traffic, by (a) minimizing intrusion on the adjacent lane while ensuring sufficient clearance for safety, and (b) having the ability to abort the overtaking maneuver if safety is compromised, e.g., due to the oncoming traffic, which is illustrated in Fig. \ref{intermediaterefselection}.

We propose an autonomous overtaking method that uses rules to choose an appropriate behaviour (maneuver) at each time using a finite state machine (FSM). 
For each behaviour, safe and reachable intermediate points are generated, inspired by \cite{dixitTrajectoryPlanningTracking2018}. 
Feasible and collision-free trajectories are then planned using model predictive control (MPC) with non-linear obstacle constraints. 
The same system components are used for all maneuvers, including staying on a lane, following a leading vehicle, overtaking, and aborting the overtake.
The method is experimentally evaluated in a high-fidelity simulation.  

The primary contributions of this work are:
\begin{enumerate}
    \item A novel autonomous overtaking method that
    \begin{enumerate}
        \item minimizes intrusions on to adjacent lane.
        \item has the ability to abort and merge back if required.
        \item can handle various lane and obstacle configurations within the same approach.
    \end{enumerate}
    \item Experimental evaluation of the method using a high-fidelity simulation environment to demonstrate that
    \begin{enumerate}
        \item the generated trajectories are feasible and respect safety constraints.
        \item the same trajectory planning method is able to handle multiple driving behaviors.
    \end{enumerate}
\end{enumerate}

\section{Related Works}

\begin{figure}[!t]
    \includegraphics[width=\linewidth]{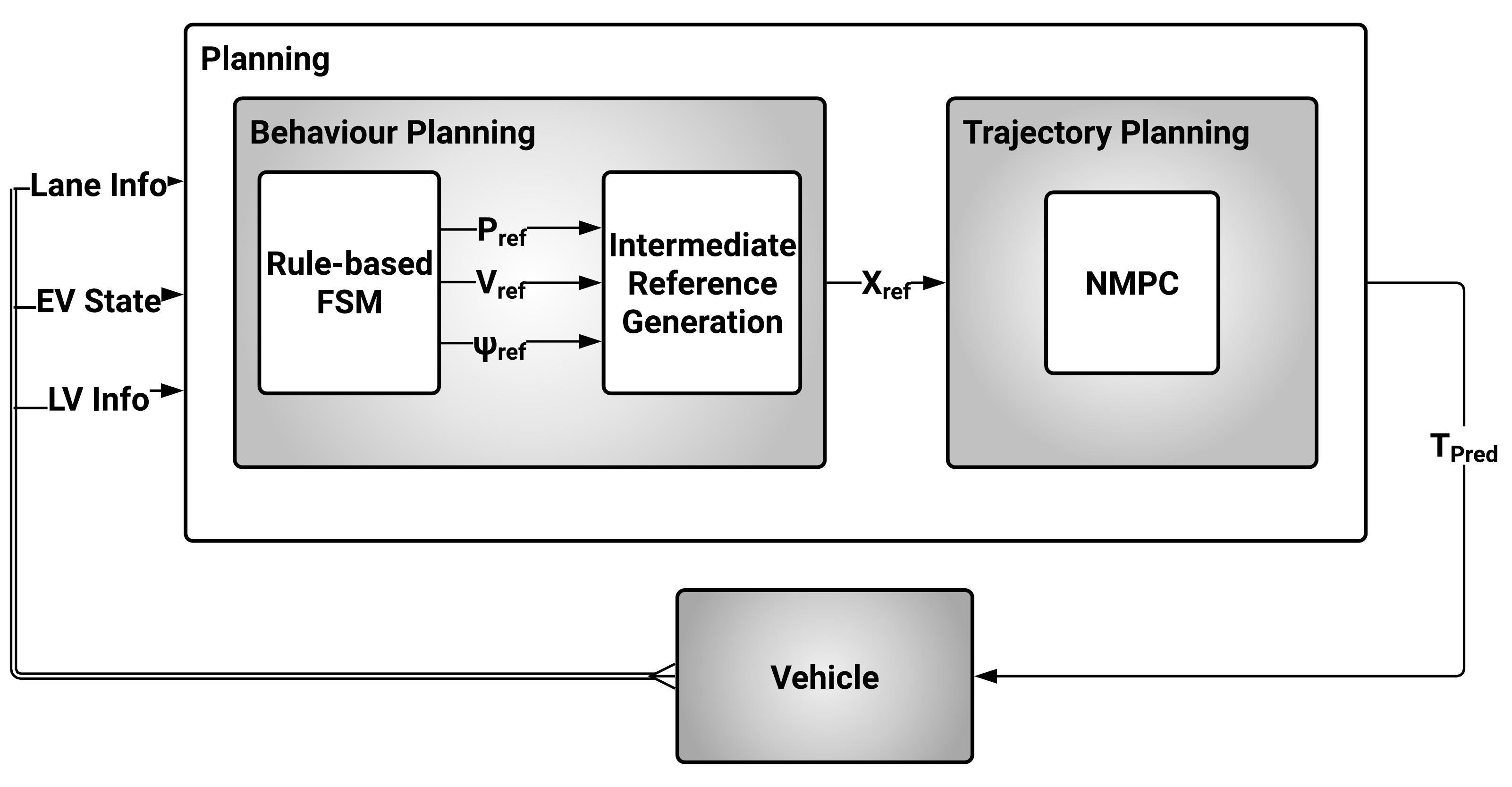}
    \centering
    \caption{Proposed behaviour and trajectory planning method block scheme.}   
    \label{proposed_control_architecture}
\end{figure}

Overtaking maneuvers consist of two decision problems: behavior choice (when to overtake, whether to abort) and trajectory planning (which trajectory to use for a particular behavior). 
These problems are typically addressed separately.
Exceptions can be found in some recent works that propose end-to-end solutions using neural network based controllers trained using reinforcement learning \cite{liqiu2020,yuluyang2020,liuPredictiveFreewayOvertaking2019}. 
However, the development of end-to-end reinforcement learning methods is still in its infancy and the methods lack stability, robustness, or optimality.

Established techniques for behavior choice include FSMs \cite{buehlerDARPAUrbanChallenge2009a} and Markov decision processes \cite{brechtelProbabilisticDecisionmakingUncertainty2014a}. 
In this work, we follow the established idea of FSMs for the behavior choice.

Vehicle trajectories can be planned e.g. by sampling based planners such as Rapidly-exploring Random Trees \cite{khaksarApplicationSamplingBasedMotion2016}, adaptive control \cite{petrov2014}, fuzzy logic \cite{armagan2020}, or MPC \cite{molinariEfficientMixedInteger2017}.
Paden \textit{et al.} \cite{padenSurveyMotionPlanning2016} and  Dixit \textit{et al.} \cite{dixitTrajectoryPlanningAutonomous2018a} provide extensive surveys on the topic. 

MPC, which we also apply in this work, is a popular choice for trajectory planning in autonomous overtaking for the reason that it can incorporate both vehicle dynamics constraints as well as obstacle constraints.
Many existing works impose some limitations in order to simplify the planning and decrease computational costs.
For example, the solution space can be restricted as in Li \textit{et al.} \cite{liDynamicTrajectoryPlanning2019}, which proposed a non-linear MPC (NMPC) to optimize lateral movement according to road conditions. 
Restricting the optimization to lateral motions prevents the approach to be used in cases with dynamic obstacles. For this reason, in this work we optimize both lateral and longitudinal components of the trajectory.

The computational cost of MPC solutions can also be reduced by limiting the planning horizon. In our work, we follow the idea proposed by Dixit \textit{et al.} \cite{dixitTrajectoryPlanningTracking2018} to generate intermediate planning targets using risk maps and reachability, which allows reducing the planning horizon and computation. 

A central issue in MPC formulations is the choice of a dynamics model. Some works, such as 
Murgovski \textit{et al.} \cite{murgovskiPredictiveCruiseControl2015} and Molinari \textit{et al.} \cite{molinariEfficientMixedInteger2017}, model the vehicle as a point mass, which does not take the non-holonomic kinematics of typical cars into account. However, the computational cost of solving MPC with complete dynamics models may be prohibitive. For that reason, we follow the compromise of using a bicycle model for dynamics, which accounts for the non-holonomic kinematics of a car, similar to \cite{dixitTrajectoryPlanningTracking2018}.

Several possible formulations for obstacle constraints exist. 
The simplest solution is to use a single linear constraint that varies during the overtake \cite{dixitTrajectoryPlanningTracking2018}. 
While this simplifies the MPC, this formulation has the challenge that it cannot account for complex scenarios with potentially multiple vehicles since the single linear is not sufficient to represent them.
Rectangular constraints for obstacle regions can also be used \cite{molinariEfficientMixedInteger2017} leading to mixed integer formulations. 
In this paper, we extend previous works by using nonlinear functions to represent obstacle boundaries.

While environment perception is an essential component of AVs, autonomous overtaking is typically considered primarily from a decision making point of view, such that the perception problems are considered to be solved, similar to this paper. Specific perception related problems, such as lack of information and unforeseen circumstances, were recently considered by \cite{andersenTrajectoryOptimizationSituational2020}, but the work did not take into account overtaking of moving vehicles. 
Beyond direct sensor-based perception, autonomous overtaking has also been studied in the context of networked vehicles, where the vehicle-to-vehicle communication is used to resolve the perception challenges
\cite{lattarulo2021}.

To summarize, the work presented in this paper goes beyond state-of-the-art by (i) including an abort maneuver in the overtaking; (ii) proposing non-linear obstacle constraints; and (iii) optimizing the trajectory, such that intrusion on adjacent lane is minimized.

\section{Proposed Method}
\label{section:proposedmethod}

    \begin{algorithm}[!t]
    \small
    \DontPrintSemicolon
    \While{TRUE}{\label{InRes1}
    \Begin(Behaviour Planning Phase){
    $S \longleftarrow generateSafeSet()$\;\label{algo:safe}
    $R \longleftarrow generateReachableSet()$\;\label{algo:reach}
    $S_{SR} = S\cap R$\;\label{algo:safereach}
    $\textbf{\text{p}}_{ref},v_{ref},\psi_{ref} \longleftarrow selectBehaviour()$ \;\label{algo:behaviour}
    $\textbf{\text{x}}_{ref} \longleftarrow getIntermediateRef()$ \;\label{algo:intermediate}
    }
    \Begin(Trajectory Planning Phase){
    $\textbf{\text{x}} \longleftarrow getCurrentState() $ \;\label{algo:nmpcstart}
    $\bm{g}_i \longleftarrow getCollisionAvoidanceConstraints()$ \;
    $\textbf{\text{u}} \longleftarrow  solveNMPC(\textbf{\text{x}},\textbf{\text{x}}_{ref},\bm{g}_i)$ \;\label{algo:nmpcstop}
    }
    }
    \caption{Planning Algorithm. \label{planningalgorithm}}
\end{algorithm}

\subsection{Method Overview}
The architecture of the proposed planning method presented in Fig. \ref{proposed_control_architecture} is broadly split into two modules: behaviour and trajectory planning. 
The overall planning algorithm is described in Algorithm \ref{planningalgorithm}. 
The behaviour planning module (Lines \ref{algo:safe} - \ref{algo:intermediate}) first identifies safe and reachable regions and makes the decision on required maneuver (i.e., lane keeping, overtaking or aborting) using an FSM based on heuristic rules. 
Next, a suitable reference target that conforms with the current maneuver is generated. 
To decrease the planning horizon, an intermediate reference state that is safe and reachable is then selected. The trajectory planning uses then non-linear MPC to plan an optimal trajectory to the intermediate reference target (Lines \ref{algo:nmpcstart} - \ref{algo:nmpcstop}). \\
\textit{Remark 1:} the proposed solution assumes that (Figure \ref{fig:reachable_safe_set}):
\begin{itemize}
    \item The state of the Ego Vehicle (EV) is fully observable (i.e., pose, velocity and acceleration with respect to World Frame (W-Frame) are available).
	\item The vehicle is able to detect and extract information about road features (i.e., lane edges and centers, road boundaries) and other traffic participants (i.e., location, velocity, heading) in 20 m radius.
\end{itemize}

\subsection{Behaviour Planning}
\label{section:behaviourplanning}
The behaviour planning consists of three steps. Initially, the safe areas devoid of the other vehicles keeping in mind the relative/absolute velocities and conforming to lane boundaries around the EV is identified resulting in a set of safe points (i.e., safe set). An appropriate maneuver is then selected along with a corresponding final reference pose and velocity associated with the maneuver. This information in conjunction with reachable areas is used to generate an intermediate reference target $\textbf{\text{x}}_{ref}$ for the trajectory planning.\

\subsubsection{Determining Safe and Reachable Set}
\label{section:safeReachSet}

\begin{figure}[!t]
    \centering
    \includegraphics[width=\linewidth]{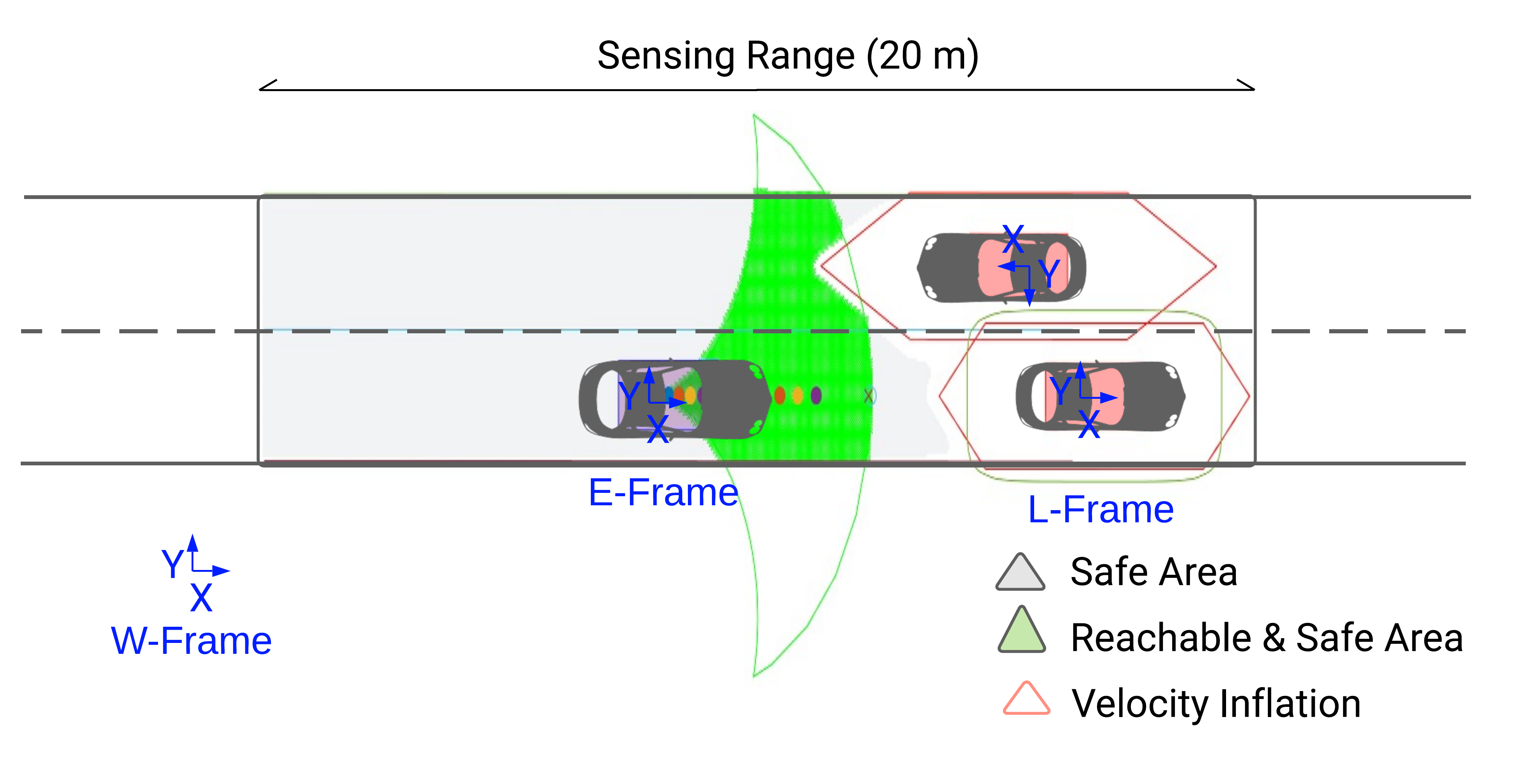}
    \caption{The safe and reachable sets illustration: shown in green. Gray area symbolizes drivable region.}
    \label{fig:reachable_safe_set}
\end{figure}

The safe regions of the road are represented by artificial potential fields as described in  \cite{wolfArtificialPotentialFunctions2008}. A local risk map of the surroundings is created by combining the information about lane/road and other vehicles, which can then be used to identify safe driving zones. The road/lane represented by a potential function is given by an exponential function so that the risk value approaches infinity at road/lane edges. This ensures that the road edges and space beyond are marked as unsafe.\

A repulsive potential field  \cite{rowlinsonYukawaPotential1989} is generated around each obstacle vehicle (annotated as polygons) to mark them as unsafe areas, thus, enabling the EV to keep a safe distance from them. These fields are exponential and approach infinity near the boundaries of the obstacle vehicles. Since the relative and absolute velocities of EV, LV are important factors, when overtaking or aborting, this information is encoded in the triangles, appended to the original obstacle polygon to serve as safety margins during the safe set generation. These velocity-depended triangles ensure that the EV leaves distance in-front and behind the LV when following or overtaking.\

The safe set $S$ (calculated in the EV frame or E-Frame) given by points in a set of all points in the sensing radius $G$ that have total risk values below a certain safe threshold $U_{threshold}$:
\begin{equation}
S = \{\textbf{\text{p}} \in G : U(\textbf{\text{p}}) \leqslant U_{threshold} \}
\end{equation}
where $p$ is any point in $G$ and $U(\textbf{\text{p}})$ is a combined potential field that is restricted by obstacle vehicles and road boundaries.\

The reachable set ($R$) defines all points the EV can reach in the entire time horizon \textit{$T_h$} (planning horizon). The reachable set $R \subset \mathbb{R}^{2}$ for time \textit{$T_h$} can be represented using polytopes that bound the reachable space on the road. The bicycle kinematic model \cite{kongKinematicDynamicVehicle2015a} is utilized to find these bounds. Using extremes of actuation, namely, front steering angle \textit{$\delta_f$}, and desired/reference velocity $v_{ref}$, the system plans for time \textit{$T_h$}. Boundary of reachable set $R \subset \mathbb{R}^{2}$ is given by:
\begin{equation} \label{reachble1}
\delta _{fmin}<\delta _{f}<\delta _{fmax} ;\
v =  v_{ref};\
a_x \leq 0,
\end{equation}
where $v$ and $a_x$ are current longitudinal velocity and acceleration of the EV, respectively.\

Both $S$ and $R$ are updated at each time step so that the risk map reflects the dynamic changes in the environment. The final safe and reachable set $S_{SR}$ is obtained from the intersection of safe set and reachable set (Fig. \ref{fig:reachable_safe_set}) as follows:
\begin{equation} \label{Fpot}
S_{SR} = S\cap R.
\end{equation}

\subsubsection{Rule-Based Finite State Machines}

\label{section:FSM}
\begin{figure}[!t]
    \includegraphics[width=\linewidth]{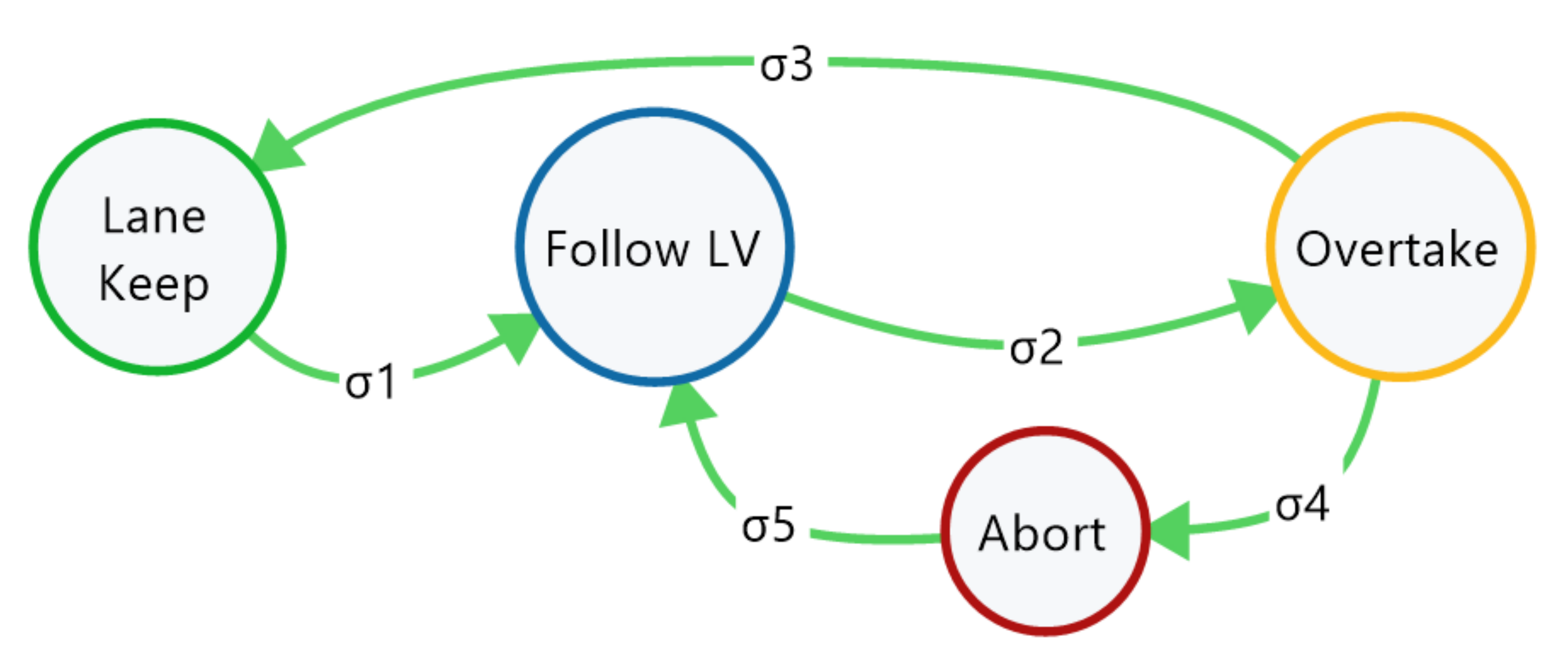}
    \centering
    \caption{The planner finite state machine}   
    \label{flowchart}
\end{figure}

The approach for the selection of the maneuver was kept simple by using an FSM since this study primarily focuses on the planning of overtaking and aborting maneuvers. The FSM is easy to implement, and the technique is efficient in deterministic decision-making. The planner state machine $M$ (Fig. \ref{flowchart}) is written as:
\begin{equation}
M = (H, \Sigma, \delta, s_0), 
\end{equation}
where $H= \{L, F, O, A\}$ is the set of states corresponding to maneuvers, $\Sigma$ is the set of inputs symbols corresponding to perceptual events triggering state transitions, $\delta$ is the state transition function $H  \times \Sigma \rightarrow H$, and $s_0\in H$ is the initial state.\

Each state corresponds to the maneuver that determine the reference pose $\textbf{\text{p}}_{ref}$ and velocity $v_{ref}$ which is eventually an input to the trajectory planning. All reference variables are generated with respect to E-frame. The pose $\textbf{\text{p}}_{ref}$ are chosen relative to the velocity-depended triangle vertex so as to automatically adapt to the different relative/absolute velocities of the EV and LV. The states $H$ are defined as follows:
\begin{itemize}
    \item \textbf{L} - Lane Keeping: This is achieved by selecting a desired pose $\textbf{\text{p}}_{ref}$ that is distance $d_{lanekeep}$ away from the current position of the EV and in the center of the lane. $v_{ref}$ is the desired set velocity during cruising.
    \item \textbf{F} - Follow Lead Vehicle: This is achieved by selecting a desired pose $\textbf{\text{p}}_{ref}$ corresponding to the vertex of velocity triangle behind the LV and in the center of the lane. $v_{ref}$ is the set as the velocity of the LV.
    \item \textbf{O} - Overtaking: The desired pose $\textbf{\text{p}}_{ref}$ is selected as the $d_{SafeOvertakeZone}$ distance from vertex of velocity triangle in front of the LV to facilitate overtaking. The desired velocity is set as $v_{ref} > v_{LV}$ to facilitate overtaking. 
    \item \textbf{A} - Aborting: During aborting, the vertex of velocity triangle behind the LV is chosen as the desired pose $\textbf{\text{p}}_{ref}$ and velocity is set as $v_{ref} < v_{LV}$ to facilitate fall back and merge back to the lane.  
\end{itemize}

State transitions are triggered by perceptual events corresponding to input symbols $\Sigma = \{\sigma_1, \sigma_2, \sigma_3,  \sigma_4, \sigma_5\}$. 
The events are defined using heuristic rules based on the current poses and velocities of EV and LV, thus, replicating a human decision making process. 
The transitions are triggered as follows: 
\begin{itemize}
    \item \textbf{$\sigma_1$} - when the EV detects a vehicle less than a fixed distance away in its sensing range.
    \item \textbf{$\sigma_2$} - when situations are favourable for overtaking, or it is manually requested.
    \item \textbf{$\sigma_3$} - when overtaking is successfully completed such that the EV has surpassed the velocity vertex in front of the LV. 
    \item \textbf{$\sigma_4$} - when potential collision before completion of the overtake is predicted, or abort of the overtake is manually triggered.
    \item \textbf{$\sigma_5$} - the aborting of the overtake maneuver is complete such that the EV has merged back to the lane behind the LV.
\end{itemize}
Even though simplified in this context, the problem of initiating these transitions (especially $\sigma_2$ and $\sigma_4$) is not trivial and depends on various factors like perception limitations, uncertainties in measurements and estimation, and, risk estimations like time to colllision.

\subsubsection{Intermediate Reference Generation}
\label{section:intermediateRefGeneration}

An intermediate reference target is chosen from the reachable and safe areas (Fig. \ref{fig:reachable_safe_set}). This reduces an overall complexity and helps in planning feasible trajectories. Moreover, the safe regions are already available as a set of points $S$.\

A suitable intermediate reference state $\textbf{\text{x}}_{ref}$ is selected from $S_{SR}$, corresponding to maneuver and the reference pose and velocity selected by the behaviour. This is achieved by selecting an intermediate point $\textbf{\text{p}}_{interim}$ on the road that belongs to $S_{SR}$ and minimizes the distance to the final desired position $\textbf{\text{p}}_{ref}$: 
\begin{equation} \label{intermediatepref}
    \textbf{\text{p}}_{interim} = \argmin_{\textbf{\text{p}}\in S_{SR}}(\| \textbf{\text{p}}-\textbf{\text{p}}_{ref}\|_2).
\end{equation}

\begin{equation}
    \textbf{\text{x}}_{ref} = [\textbf{\text{p}}_{interim}, \psi_{ref}, v_{ref}]    
\end{equation}

This process is repeated at every time step, after which $\textbf{\text{p}}_{interim}$ moves closer to final target $\textbf{\text{p}}_{ref}$ until it finally coincides with it. The intermediate reference point selection process remains the same regardless of the selected maneuver. This iterative process guides the NMPC to perform the required maneuver, consequently reducing design complexity. It also results in minimal intrusion onto adjacent lanes while overtaking. Furthermore, it captures any sudden changes in the environment or in the state of the LV, thus, ensuring that $\textbf{\text{x}}_{ref}$ is always safe, i.e., $\textbf{\text{p}}_{ref} \in S_{SR}$.

The process of intermediate target selection is illustrated in Fig. \ref{intermediaterefselection}.
Finally, the intermediate reference target $\textbf{\text{x}}_{ref}$ is provided as an input reference to the NMPC.

\subsection{Trajectory Planning}
\label{section:trajectoryplanning}
The next phase of planning generates a trajectory over a finite horizon ($N$) that follows the system dynamics and devoid of collisions, obeys state and control limits. Since this is a constrained finite-time optimal control problem, it was formulated as an NMPC as:
\begin{equation}
\label{NMPC_Formulation}
\begin{aligned}
\underset{\textbf{\text{x}}_1, ..., \textbf{\text{x}}_{N},\textbf{\text{u}}_0, ..., \textbf{\text{u}}_{N-1}}{\text{min}} \quad & \ell_f(\textbf{\text{x}}_N)+\sum_{k=0}^{N-1}\ell(\textbf{\text{x}}_k, \textbf{\text{u}}_k) \\
\text{subject to} \quad & \textbf{\text{x}}_{k+1} = f(\textbf{\text{x}}_k, \textbf{\text{u}}_k), \\ 
& \bm{g}(\textbf{\text{x}}_k)\geq \textbf{\text{0}}, \\
& \textbf{\text{x}}_{min} \leq \textbf{\text{x}}_k \leq \textbf{\text{x}}_{max}, \\
& \textbf{\text{u}}_{min} \leq \textbf{\text{u}}_k \leq \textbf{\text{u}}_{max}, \\
& \textbf{\text{x}}_0 = \bar{\textbf{\text{x}}},
\end{aligned}
\end{equation}
for all $k=0,..., N-1$. $\textbf{\text{x}}_k = [x_k, y_k, \psi_k, v_k]$ is the state vector including longitudinal and lateral positions of the center of mass in an inertial frame (W), inertial heading angle and linear speed of the vehicle, respectively. The control input vector $\textbf{\text{u}}_k = [a_k, \delta_k]$ involves acceleration and steering angle. $\ell_N$ and $\ell$ are the final and the running cost functions, which were selected as follows:
\begin{equation}
\label{Cost_Functions}
\begin{aligned}
\ell_N(\textbf{\text{x}}_N) = & (\textbf{\text{x}}_{ref}-\textbf{\text{x}}_N)^{\top}Q_N(\textbf{\text{x}}_{ref}-\textbf{\text{x}}_N) \\
\ell(\textbf{\text{x}}_k, \textbf{\text{u}}_k) = & (\textbf{\text{x}}_{ref}-\textbf{\text{x}}_k)^{\top}Q_k(\textbf{\text{x}}_{ref}-\textbf{\text{x}}_k) + \textbf{\text{u}}_k^{\top}R_k\textbf{\text{u}}_k,
\end{aligned}
\end{equation}
$f(\textbf{\text{x}}_k, \textbf{\text{u}}_k)$ is the function of state transition dynamics where it follows the nonlinear bicycle model described in \cite{kongKinematicDynamicVehicle2015a}. Equations of motion for the model can be written as:
\begin{equation} \label{sysmodel1}
\dot{x} = v\cos (\psi + \beta)
\end{equation}
\begin{equation} \label{sysmodel2}
\dot{y} = v\sin (\psi + \beta)
\end{equation}
\begin{equation} \label{sysmodel3}
\dot{\psi} = \frac{v}{l_f+l_r} \cos (\beta) \tan(\delta )
\end{equation}
\begin{equation} \label{sysmodel4}
\dot{v} = a
\end{equation}
\begin{equation} \label{sysmodel5}
\beta = \tan^{-1}\left (\frac{l_r}{l_f+l_r} \tan(\delta)\right )
\end{equation}
where $\beta$ is the angle of the current velocity of the center of mass with respect to the longitudinal axis of the car. \textit{$l_r$} and \textit{$l_f$} are the distances from the center of the mass (CoM) of the vehicle to the front and rear axles, respectively.

Even though the $\textbf{\text{x}}_{ref}$ provided by Behaviour Planning Phase (Section \ref{section:intermediateRefGeneration}) is safe, it is necessary to ensure that the trajectory planned by the NMPC is also safe and free from collisions. To achieve this, the obstacle avoidance is expressed as state constraints in the form of constraint function $\bm{g}(\textbf{\text{x}}_k)$ where the obstacles are represented using higher order ellipses. This ensures that the admissible set of possible solutions for NMPC is not limited due to over-fitting required for inscribing rectangular obstacle boundaries inside simple ellipses. Thus the constraint function is formulated as:
\begin{equation}
\label{Constraint_Function}
\begin{aligned}
\bm{g}(\textbf{\text{x}}_k) = & \Bigg(\frac{(x_k-x_{e,i})\cos(\phi_{i})-(y_k-y_{e,i})\sin(\phi_{i})}{a_{i}}\Bigg)^n\\
+& \Bigg(\frac{(x_k-x_{e,i})\sin(\phi_{i})-(y_k-y_{e,i})\cos(\phi_{i})}{b_{i}}\Bigg)^n-1,
\end{aligned}
\end{equation}
where $i$ indicates the index of the obstacle.

The parameters ($x_e, y_e, a, b, \phi, n$) of the ellipse encompassing the rectangular obstacles, i.e., other vehicles, are calculated based on the pose and the dimensions of the obstacle vehicle. The ellipse is also padded with the length and breadth of the EV, and an inflation factor $\alpha$ to ensure no collisions. The velocity dependent triangles used in safe set calculation is not considered in this context as it is artificial and would not result in collisions even if violated. Moreover, the way of intermediate reference point generation already takes these safety margins into account.

In addition to the obstacle avoidance constraints, the states and the control inputs are also constrained by the physical capabilities of the vehicle. Lastly, the measured current state of the vehicle ($\bar{\textbf{\text{x}}}$) initializes the $\textbf{\text{x}}_0$ and initiates the optimization through prediction horizon in NMPC.

\section{Experimental Results}

\subsection{Experimental environment}
The proposed planning framework was implemented in MathWorks' MATLAB/Simulink environment. A closed-loop simulation was developed with a 14 degrees of freedom vehicle model for the EV and other traffic participants using the Vehicle Dynamics and Automated Driving Toolboxes. The NMPC was implemented using the MPC Toolbox and a sequential quadratic programming solver \cite{blaszczykObjectLibraryAlgorithms2007} of MATLAB optimization toolbox. A modified tracking controller from MATLAB Vehicle Dynamics Toolbox was used for tracking the trajectories generated by the NMPC at each time instant.\

The parameters required by NMPC were selected as follows:
\begin{itemize}
    \item Prediction and control horizon: N = 10
    \item Time horizon: $T_h$ = 1s
    \item Weights in cost functions: \\
    $Q_{i} = diag\{0, 5, 20, 10\}\text{ for } i=0,\dots, 4  \\
    Q_{i} = diag\{0, 10, 20, 10\}\text{ for } i=5,\dots, 6 \\
    Q_{i} = diag\{0, 10, 50, 10\}\text{ for } i=7,\dots, 8 \\
    Q_{i} = diag\{0, 50, 50, 30\}\text{ for } i=9 \\
    R_{i} = diag\{5, 50\}\text{ for } i=0,\dots, 9$
\end{itemize}

The weights $Q_{i}$ were chosen so as to prioritize lateral position ($y$), velocity ($v$) and heading ($\psi$), and obtain trajectories that minimize unnecessary detours from the straight-line trajectory. Steering inputs were also penalized more than acceleration inputs in $R_{i}$ to achieve smooth trajectories devoid of unnecessary turns. $T_h$ was as a trade-off between trajectory length and computation costs.  

\subsection{Simulation Results}

\begin{figure}[!t]
  \centering
  \includegraphics[width=\linewidth]{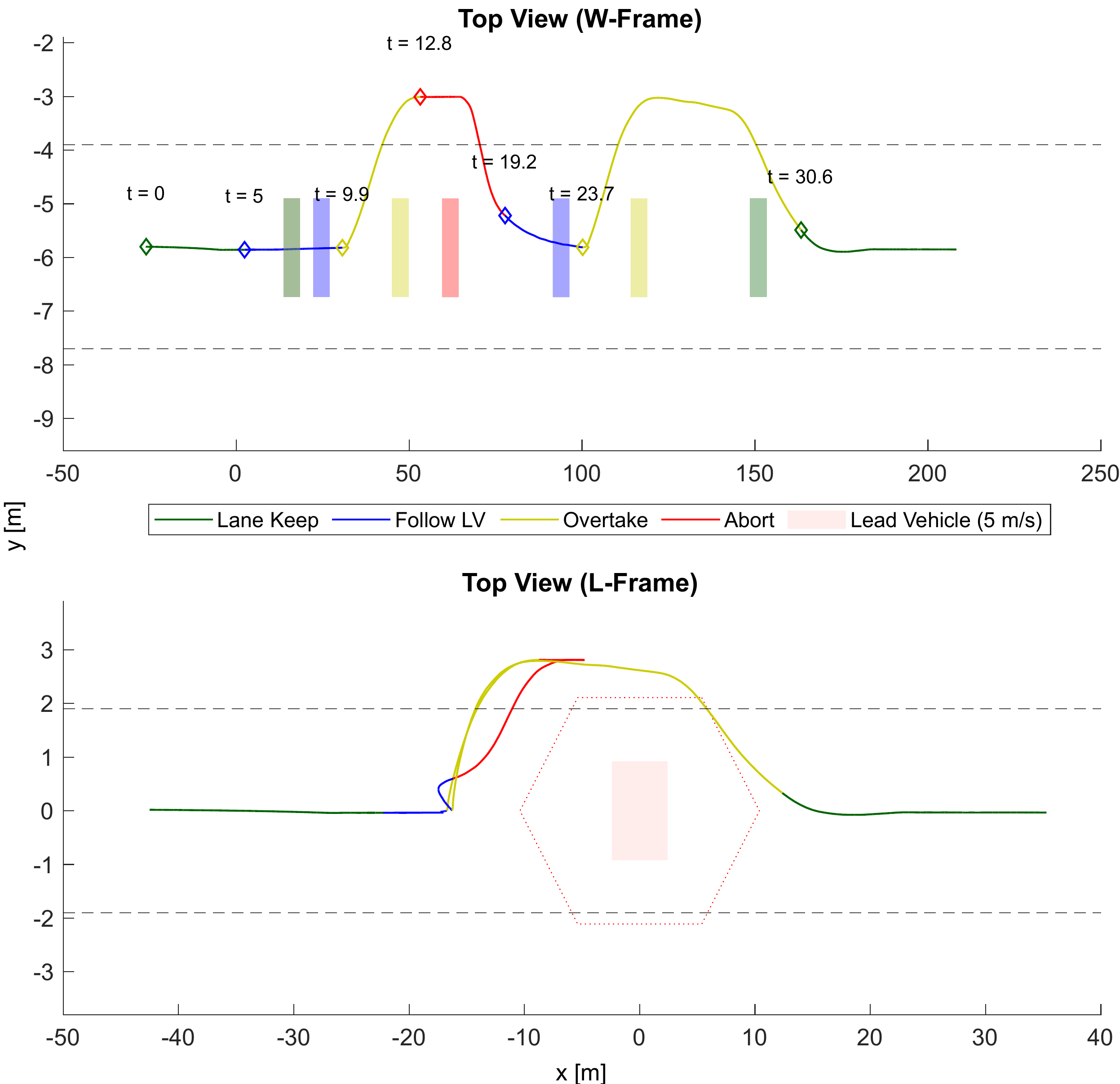}
  \caption{Actual trajectories of the EV and LV during the scenario simulation (shown in both W-Frame and L-Frame). Dashed lines indicate the lane boundaries.}
  \label{fig:topview}
\end{figure}

The scenario was set up to assess the ability of the proposed framework to successfully overtake and abort in case of a potential collision. 
As illustrated in Fig. \ref{fig:topview}, initially EV accelerates and gains desired speed while following the lane center (L). When it encounters LV in its sensing range, the EV switches to follow lead vehicle mode (F). The LV is moving at a velocity of 5 m/s. At an opportune moment, overtaking is triggered and the EV proceeds to overtake (O) the LV, but then has to abort the maneuver (A) to avoid some potentially catastrophic collision. Therefore, the EV merges back to the lane and then reattempts to overtake successfully completing the maneuver. The planned and actual trajectories of the EV were analyzed along with the control inputs to evaluate the performance.\

\begin{figure}[!t]
  \centering
  \includegraphics[width=\linewidth]{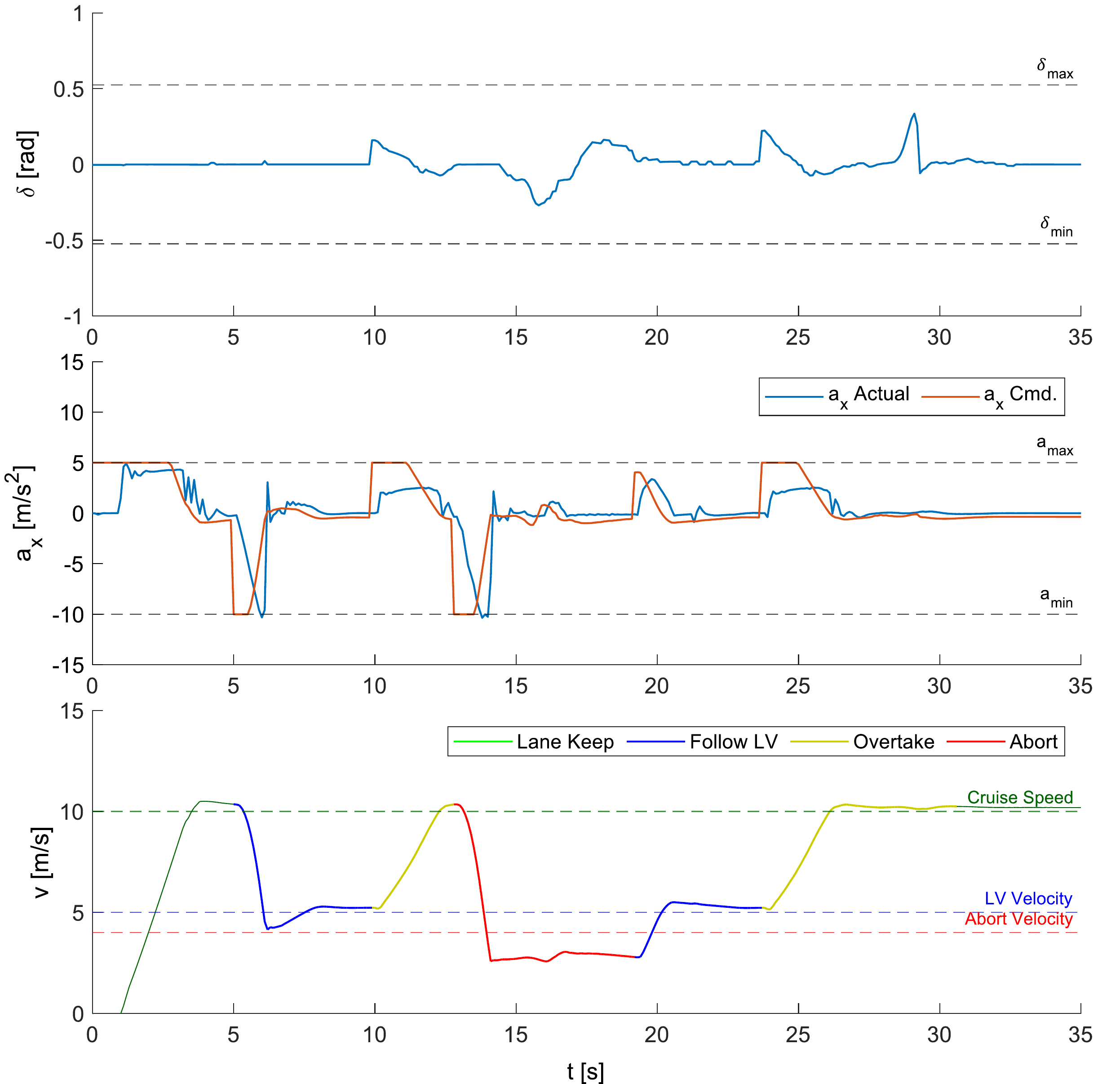}
  \caption{Optimal Acceleration and Steering commands generated by the NMPC. Dotted lines indicate the vehicle limits. The velocity curve during the whole scenario is also shown. The target velocities during each maneuver phase are indicated using dotted lines.}
  \label{fig:accsteervel}
\end{figure}

Figure \ref{fig:topview} shows the actual trajectory of the EV as it goes through the different phases of the test scenario. The trajectory is shown in both the W-frame and the inertial L-frame to gain a better picture of the path of the EV. Each maneuver is indicated using different colors, namely, Lane Keep (Green), Follow Lead Vehicle (Blue), Overtaking (Yellow), Abort (Red). The pose of the LV at the start of each maneuver phase is as a rectangle of an appropriate color.\  

At t = 0s, the EV is at rest and accelerates to attain the desired speed $v_{des}$ of 10 m/s as seen in Fig. \ref{fig:accsteervel}. When the EV detects the LV as it falls into its sensing range, the EV decelerates to match the velocity of LV (t = 5s) and follows while keeping a safe distance. At t = 9.9s overtaking is initiated. The trajectory planned during overtake only intrudes onto the adjacent lane so as to keep a safe distance from the obstacle vehicle.\

At t = 12.8s abort is initiated due to potential collision event suggesting unsafe overtake conditions. The EV rapidly decelerates to a velocity ($v_{des} < v_{LV}$) (Fig. \ref{fig:accsteervel}). The trajectory planned guides the EV to safely merge back to lane again while keeping a safe distance from the LV. After a successful merge, the EV again switches to follow mode. At t = 23.7s, the overtake is again attempted. As no obstacles or potential collisions are detected, the EV successfully completes the overtaking maneuver. When the EV is at a safe distance in front of the LV it merges to the lane and resumes lane keeping mode (t = 30.6s).\

\begin{figure}[!t]
  \centering
  \includegraphics[width=\linewidth]{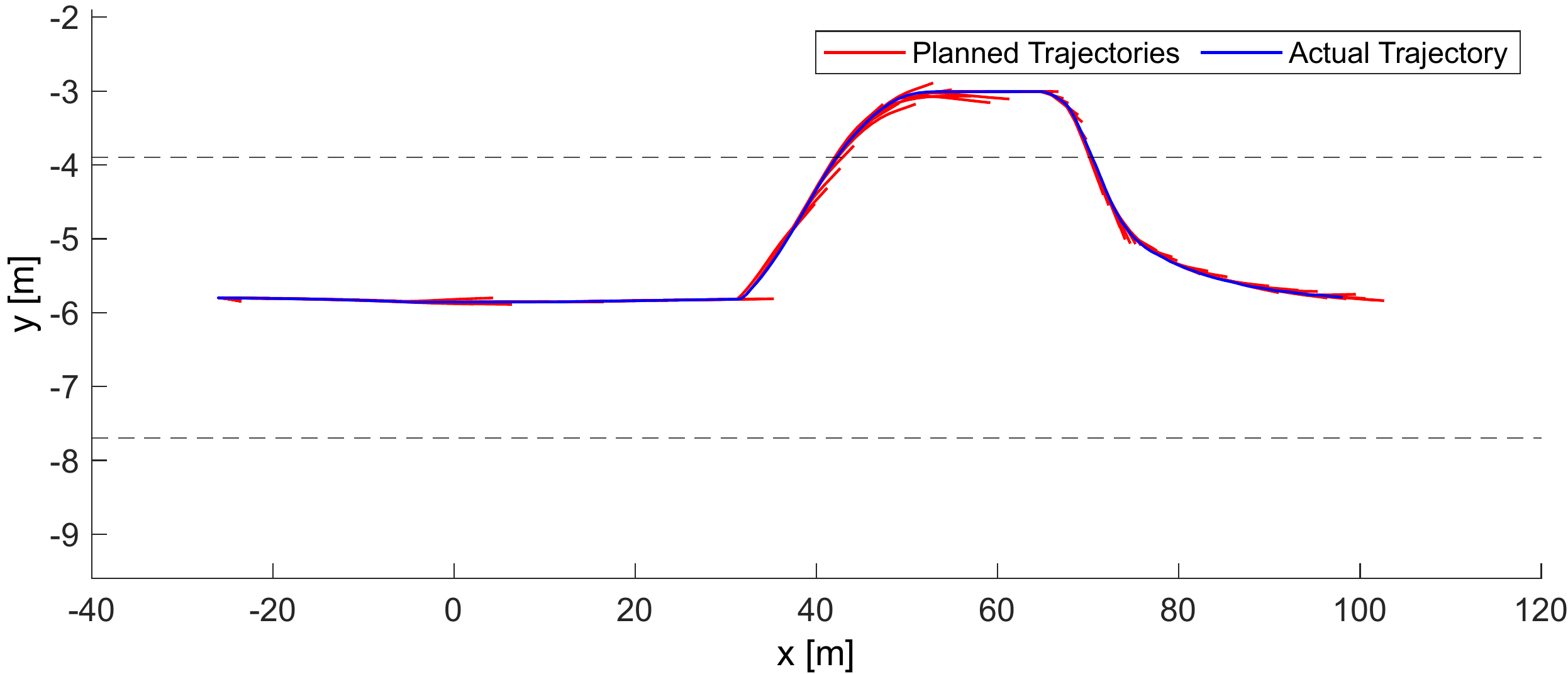}
  \caption{Comparison of  actual trajectory vs predicted trajectories at each time step (in W-Frame). Dashed lines indicate the lane boundaries.}
  \label{fig:xpred}
\end{figure}

As seen in Fig. \ref{fig:topview}, the trajectories of the EV obey the obstacle avoidance constraints and hence are safe. The trajectories also do not command unreasonable acceleration ($a$) or steering commands ($\delta$) and are well within limits imposed (Fig. \ref{fig:topview}). In Fig. \ref{fig:xpred} the planned trajectories versus the actual trajectory of the EV are demonstrated. The short prediction horizons enabled by the intermediate reference target selection allow for dynamic re-planning of the trajectory to account for the model mismatch and changes in the environment.

The scenario was repeated for different velocities of the EV and the LV (up to 20 m/s), with similar results. The EV was able to successfully overtake and abort the maneuver if desired in all cases without modifications of any parameters. This demonstrate the wide range of applicability of the proposed method under variable driving conditions.

\section{Conclusions}

In this paper, we presented a method for autonomous overtaking, which allows the overtaking to be aborted if safety is compromised. 
The ability to abort makes the system more reactive, and thus increases the range of situations, where overtaking can be performed safely.
Our work demonstrates that by splitting the overall problem into behaviour and trajectory planning the same framework is able to handle various driving behaviors, and that new behaviors can be constructed easily without changing the trajectory planner.
The use of an MPC-based trajectory planner allows the method to handle simultaneously multiple objectives, enforcing collision-free trajectories and minimizing intrusion onto the adjacent lane, while retaining guarantees on stability and constraint satisfaction.  

A simple FSM was adopted in our work for behaviour planning, similar to most other works. 
However, in complex driving environments such a solution would most likely be insufficient, because it would be infeasible to design manually an FSM that would account for all eventualities of the complex environment, while 
simpler FSMs and related rule sets would lead to either sacrificing performance due to over-conservatism or sacrificing safety due to over-optimism. 
At the moment, we are not aware of a solution to the behavior planning in complex environments that would not be prone to these sacrifices and would still be computationally feasible.
As more real-world data become available for the development of AVs, we foresee that this open problem likely can be solved by integrating machine learning as part of the solution, in order to handle the associated large state spaces.
Thus, behaviour planning appears to be an essential topic of study to realize human-level autonomous overtaking capabilities in autonomous cars.


\section*{Acknowledgment}
The authors would like to thank Dr. Shilp Dixit (ARRIVAL LTD.) for his guidance during the initial phases of the work through correspondence and being an inspiration for this paper.

\bibliographystyle{IEEEtran}
\bibliography{IEEEabrv,ITSC_2021.bib}

\end{document}